\documentclass[11pt,a4paper]{article}
\usepackage[hyperref]{emnlp2018}
\usepackage{times}
\usepackage{latexsym}
\usepackage{amsfonts}
\usepackage{amsmath}
\usepackage{algorithmic}
\usepackage{algorithm}

\usepackage[pdftex]{graphicx}
\usepackage{CJKutf8}

\usepackage{subcaption}

\usepackage{url}

\aclfinalcopy 
\def\aclpaperid{45} 

\title{An Operation Sequence Model for Explainable Neural Machine Translation}

\author{First Author \\
  Affiliation / Address line 1 \\
  Affiliation / Address line 2 \\
  Affiliation / Address line 3 \\
  {\tt email@domain} \\\And
  Second Author \\
  Affiliation / Address line 1 \\
  Affiliation / Address line 2 \\
  Affiliation / Address line 3 \\
  {\tt email@domain} \\}

\author{Felix Stahlberg \and Danielle Saunders \and Bill Byrne\\
         Department of Engineering\\
         University of Cambridge, UK \\
         {\tt  \{fs439,ds636,wjb31\}@cam.ac.uk}}

\date{}

\begin{document}
\maketitle
\begin{abstract}
We propose to achieve explainable neural machine translation (NMT) by changing the output representation to explain itself. We present a novel approach to NMT which generates the target sentence by monotonically walking through the source sentence. Word reordering is modeled by operations which allow setting markers in the target sentence and move a target-side write head between those markers. In contrast to many modern neural models, our system emits explicit word alignment information which is often crucial to practical machine translation as it improves explainability. Our technique can outperform a plain text system in terms of BLEU score under the recent Transformer architecture on Japanese-English and Portuguese-English, and is within 0.5 BLEU difference on Spanish-English.

\end{abstract}

\section{Introduction}

Neural machine translation (NMT) models~\citep{sutskever,bahdanau,convs2s,transformer} are remarkably effective in modelling the distribution over target sentences conditioned on the source sentence, and yield superior translation performance compared to traditional statistical machine translation (SMT) on many language pairs. However, it is often difficult to extract a comprehensible explanation for the predictions of these models as information in the network is represented by real-valued vectors or matrices~\citep{visual-nmt}. In contrast, the translation process in SMT is `transparent' as it can identify the source word which caused a target word through word alignment. Most NMT models do not use the concept of word alignment. It is tempting to interpret encoder-decoder attention matrices~\citep{bahdanau} in neural models as (soft) alignments, but previous work has found that the attention weights in NMT are often erratic~\citep{nmt-attention-sucks} and differ significantly from traditional word alignments~\citep{nmt-challenges,nmt-where-attention}. We will discuss the difference between attention and alignment in detail in Sec.~\ref{sec:att}. The goal of this paper is explainable NMT by developing a transparent translation process for neural models. Our approach does not change the neural architecture, but represents the translation together with its alignment as a linear sequence of operations. The neural model predicts this operation sequence, and thus simultaneously generates a translation and an explanation for it in terms of alignments from the target words to the source words that generate them. The operation sequence is ``self-explanatory''; it does not explain an underlying NMT system but is rather a single representation produced by the NMT system that can be used to generate translations along with an accompanying explanatory alignment to the source sentence. We report competitive results of our method on Spanish-English, Portuguese-English, and Japanese-English, with the benefit of producing hard alignments for better interpretability. We discuss the theoretical connection between our approach and hierarchical SMT~\citep{hiero} by showing that an operation sequence can be seen as a derivation in a formal grammar.

\section{A Neural Operation Sequence Model}
\label{sec:osnmt}

Our operation sequence neural machine translation (OSNMT) model is inspired by the operation sequence model for SMT~\citep{osm}, but changes the set of operations to be more appropriate for neural sequence models. OSNMT is not restricted to a particular architecture, i.e.\ any seq2seq model such as RNN-based, convolutional, or self-attention-based models~\citep{bahdanau,transformer,convs2s} could be used.
In this paper, we use the recent Transformer model architecture~\citep{transformer} in all experiments.

In OSNMT, the neural seq2seq model learns to produce a sequence of operations. An OSNMT operation sequence describes a translation (the `compiled' target sentence) and explains each target token with a hard link into the source sentence. OSNMT keeps track of the positions of a source-side read head and a target-side write head. The read head monotonically walks through the source sentence, whereas the position of the write head can be moved from marker to marker in the target sentence. OSNMT defines the following operations to control head positions and produce output words.

\begin{itemize}
\item \texttt{POP\_SRC}: Move the read head right by one token.
\item \texttt{SET\_MARKER}: Insert a marker symbol into the target sentence at the position of the write head.
\item \texttt{JMP\_FWD}: Move the write head to the nearest marker right of the current head position in the target sentence.
\item \texttt{JMP\_BWD}: Move the write head to the nearest marker left of the current head position in the target sentence.
\item \texttt{INSERT}($t$): Insert a target token $t$ into the target sentence at the position of the write head.
\end{itemize}

Tab.~\ref{tab:osm-breakdown} illustrates the generation of a Japanese-English translation in detail. The neural seq2seq model is trained to produce the sequence of operations in the first column of Tab.~\ref{tab:osm-breakdown}. The initial state of the target sentence is a single marker symbol $X_1$. Generative operations like \texttt{SET\_MARKER} or \texttt{INSERT}($t$) insert a single symbol left of the current marker (highlighted). The model begins with a \texttt{SET\_MARKER} operation, which indicates that the translation of the first word in the source sentence is not at the beginning of the target sentence. Indeed, after ``translating'' the identities `2000' and `hr', in time step 6 the model jumps back to the marker $X_2$ and continues writing left of `2000'. The translation process terminates when the read head is at the end of the source sentence. The final translation in plain text can be obtained by removing all markers from the (compiled) target sentence. 

\subsection{OSNMT Represents Alignments}
\label{sec:osnmt-align}

The word alignment can be derived from the operation sequence by looking up the position of the read head for each generated target token. The alignment for the example in Tab.~\ref{tab:osm-breakdown} is shown in Fig.~\ref{fig:alignment}. Note that similarly to the IBM models~\citep{ibm} and the OSM for SMT~\citep{osm}, our OSNMT can only represent 1:$n$ alignments. Thus, each target token is aligned to exactly one source token, but a source token can generate any number of (possibly non-consecutive) target tokens.

\begin{table*}[t!]
\begin{center}
\small
\begin{tabular}{r|l|l|l|}
\cline{2-4}
 & \bf Operation & \bf Source sentence & \bf Target sentence (compiled) \\ \cline{2-4}

 &		& \begin{CJK}{UTF8}{min}\colorbox{blue!15}{2000} hr の 安定 動作 を 確認 し た\end{CJK}	& \colorbox{blue!15}{$X_1$} \\
1 & \texttt{SET\_MARKER}		& \begin{CJK}{UTF8}{min}\colorbox{blue!15}{2000} hr の 安定 動作 を 確認 し た\end{CJK}	& $X_2$ \colorbox{blue!15}{$X_1$} \\
2 & 2000		& \begin{CJK}{UTF8}{min}\colorbox{blue!15}{2000} hr の 安定 動作 を 確認 し た\end{CJK}	& $X_2$ 2000 \colorbox{blue!15}{$X_1$} \\
3 & \texttt{POP\_SRC}		& \begin{CJK}{UTF8}{min}2000 \colorbox{blue!15}{hr} の 安定 動作 を 確認 し た\end{CJK}	& $X_2$ 2000 \colorbox{blue!15}{$X_1$} \\
4 & hr		& \begin{CJK}{UTF8}{min}2000 \colorbox{blue!15}{hr} の 安定 動作 を 確認 し た\end{CJK}	& $X_2$ 2000 hr \colorbox{blue!15}{$X_1$} \\
5 & \texttt{POP\_SRC}		& \begin{CJK}{UTF8}{min}2000 hr \colorbox{blue!15}{の} 安定 動作 を 確認 し た\end{CJK}	& $X_2$ 2000 hr \colorbox{blue!15}{$X_1$} \\
6 & \texttt{JMP\_BWD}		& \begin{CJK}{UTF8}{min}2000 hr \colorbox{blue!15}{の} 安定 動作 を 確認 し た\end{CJK}	& \colorbox{blue!15}{$X_2$} 2000 hr $X_1$ \\
7 & \texttt{SET\_MARKER}		& \begin{CJK}{UTF8}{min}2000 hr \colorbox{blue!15}{の} 安定 動作 を 確認 し た\end{CJK}	& $X_3$ \colorbox{blue!15}{$X_2$} 2000 hr $X_1$ \\
8 & of		& \begin{CJK}{UTF8}{min}2000 hr \colorbox{blue!15}{の} 安定 動作 を 確認 し た\end{CJK}	& $X_3$ of \colorbox{blue!15}{$X_2$} 2000 hr $X_1$ \\
9 & \texttt{POP\_SRC}		& \begin{CJK}{UTF8}{min}2000 hr の \colorbox{blue!15}{安定} 動作 を 確認 し た\end{CJK}	& $X_3$ of \colorbox{blue!15}{$X_2$} 2000 hr $X_1$ \\
10 & \texttt{JMP\_BWD}		& \begin{CJK}{UTF8}{min}2000 hr の \colorbox{blue!15}{安定} 動作 を 確認 し た\end{CJK}	& \colorbox{blue!15}{$X_3$} of $X_2$ 2000 hr $X_1$ \\
11 & stable		& \begin{CJK}{UTF8}{min}2000 hr の \colorbox{blue!15}{安定} 動作 を 確認 し た\end{CJK}	& stable \colorbox{blue!15}{$X_3$} of $X_2$ 2000 hr $X_1$ \\
12 & \texttt{POP\_SRC}		& \begin{CJK}{UTF8}{min}2000 hr の 安定 \colorbox{blue!15}{動作} を 確認 し た\end{CJK}	& stable \colorbox{blue!15}{$X_3$} of $X_2$ 2000 hr $X_1$ \\
13 & operation	& \begin{CJK}{UTF8}{min}2000 hr の 安定 \colorbox{blue!15}{動作} を 確認 し た\end{CJK}	& stable operation \colorbox{blue!15}{$X_3$} of $X_2$ 2000 hr $X_1$ \\
14 & \texttt{POP\_SRC}		& \begin{CJK}{UTF8}{min}2000 hr の 安定 動作 \colorbox{blue!15}{を} 確認 し た\end{CJK}	& stable operation \colorbox{blue!15}{$X_3$} of $X_2$ 2000 hr $X_1$ \\
15 & \texttt{JMP\_FWD}		& \begin{CJK}{UTF8}{min}2000 hr の 安定 動作 \colorbox{blue!15}{を} 確認 し た\end{CJK}	& stable operation $X_3$ of \colorbox{blue!15}{$X_2$} 2000 hr $X_1$ \\
16 & \texttt{JMP\_FWD}		& \begin{CJK}{UTF8}{min}2000 hr の 安定 動作 \colorbox{blue!15}{を} 確認 し た\end{CJK}	& stable operation $X_3$ of $X_2$ 2000 hr \colorbox{blue!15}{$X_1$} \\
17 & was		& \begin{CJK}{UTF8}{min}2000 hr の 安定 動作 \colorbox{blue!15}{を} 確認 し た\end{CJK}	& stable operation $X_3$ of $X_2$ 2000 hr was \colorbox{blue!15}{$X_1$} \\
18 & \texttt{POP\_SRC}		& \begin{CJK}{UTF8}{min}2000 hr の 安定 動作 を \colorbox{blue!15}{確認} し た\end{CJK}	& stable operation $X_3$ of $X_2$ 2000 hr was \colorbox{blue!15}{$X_1$} \\
19 & \texttt{POP\_SRC}		& \begin{CJK}{UTF8}{min}2000 hr の 安定 動作 を 確認 \colorbox{blue!15}{し} た\end{CJK}	& stable operation $X_3$ of $X_2$ 2000 hr was \colorbox{blue!15}{$X_1$} \\
20 & confirmed	& \begin{CJK}{UTF8}{min}2000 hr の 安定 動作 を 確認 \colorbox{blue!15}{し} た\end{CJK}	& stable operation $X_3$ of $X_2$ 2000 hr was confirmed \colorbox{blue!15}{$X_1$} \\
21 & \texttt{POP\_SRC}		& \begin{CJK}{UTF8}{min}2000 hr の 安定 動作 を 確認 し \colorbox{blue!15}{た} \end{CJK}	& stable operation $X_3$ of $X_2$ 2000 hr was confirmed \colorbox{blue!15}{$X_1$} \\

\cline{2-4}
\end{tabular}
\end{center}
\caption{\label{tab:osm-breakdown} Generation of the target sentence ``stable operation of 2000 hr was confirmed'' from the source sentence ``\begin{CJK}{UTF8}{min}2000 hr の 安定 動作 を 確認 し た\end{CJK}''. The neural model produces the linear sequence of operations in the first column. The positions of the source-side read head and the target-side write head are highlighted. The marker in the target sentence produced by the $i$-th \texttt{SET\_MARKER} operation is denoted with `$X_{i+1}$'; $X_1$ is the initial marker. We denote \texttt{INSERT}($t$) operations as $t$ to simplify notation. 
}
\end{table*}

\subsection{OSNMT Represents Hierarchical Structure}
\label{sec:osnmt-hiero}

\begin{figure}[t]
\centering
\includegraphics[width=0.85\linewidth]{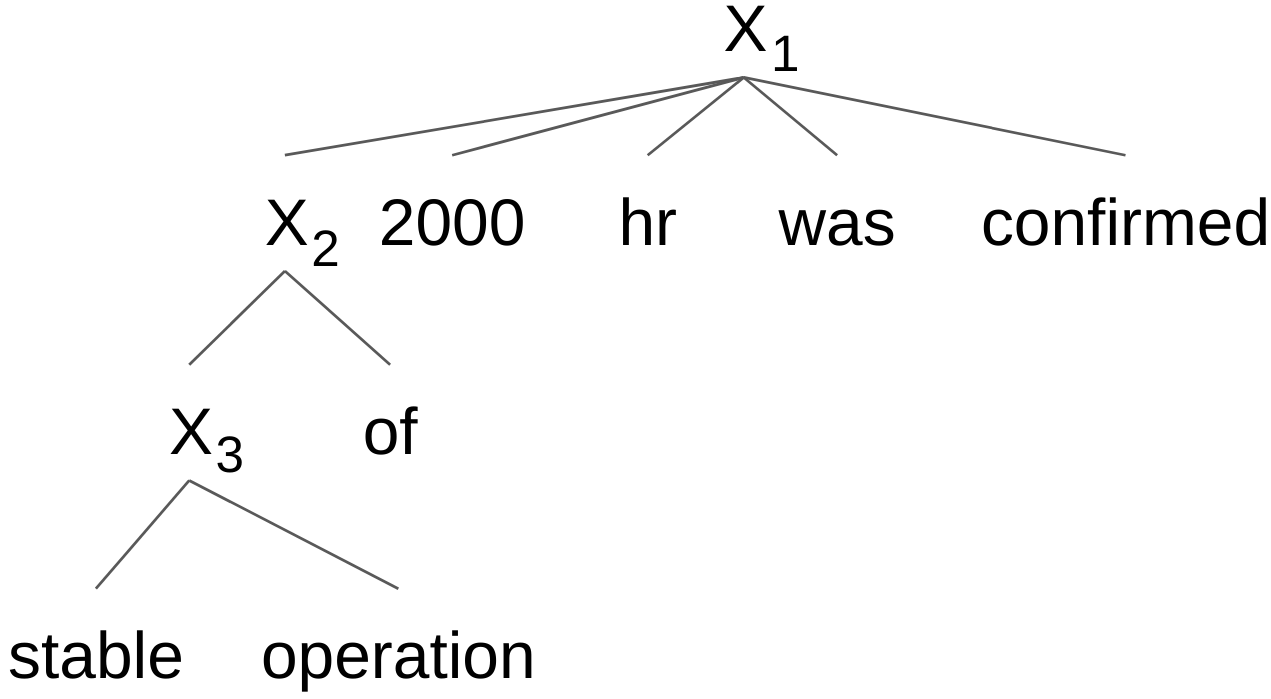}
\caption{Target-side tree representation of the operation sequence in Tab.~\ref{tab:osm-breakdown}.}
\label{fig:tree}
\end{figure}

We can also derive a tree structure from the operation sequence in Tab.~\ref{tab:osm-breakdown} (Fig.~\ref{fig:tree}) in which each marker is represented by a nonterminal node with outgoing arcs to symbols inserted at that marker. The target sentence can be read off the tree by depth-first search traversal (post-order).

More formally, synchronous context-free grammars (SCFGs) generate pairs of strings by pairing two context-free grammars. Phrase-based hierarchical SMT~\citep{hiero} uses SCFGs to model the relation between the source sentence and the target sentence. Multitext grammars (MTGs) are a generalization of SCFGs to more than two output streams~\citep{mtg,gmtg}. We find that an OSNMT sequence can be interpreted as sequence of rules of a tertiary MTG $\mathcal{G}$ which generates 1.) the source sentence, 2.) the target sentence, and 3.) the position of the target side write head. The start symbol of $\mathcal{G}$ is

\begin{equation}
[(S), (X_1), (P_1)]^\mathsf{T}
\end{equation}
which initializes the source sentence stream with a single nonterminal $S$, the target sentence with the initial marker $X_1$ and the position of the write head with 1 ($P_1$). Following \citet{gmtg} we denote rules in $\mathcal{G}$ as

\begin{equation}
[(\alpha_1),(\alpha_2),(\alpha_3)]^\mathsf{T}\rightarrow [(\beta_1),(\beta_2),(\beta_3)]^\mathsf{T}
\end{equation}
where $\alpha_1, \alpha_2,\alpha_3$ are single nonterminals or empty, $\beta_1,\beta_2,\beta_3$ are strings of terminals and nonterminals, and $\alpha_i\rightarrow\beta_i$ for all $i\in\{1,2,3\}$ with nonempty $\alpha_i$ are the rewriting rules for each of the three individual components which need to be applied simultaneously. \texttt{POP\_SRC} extends the source sentence prefix in the first stream by one token.

\begin{figure}[!t]
\centering
\includegraphics[width=0.9\linewidth]{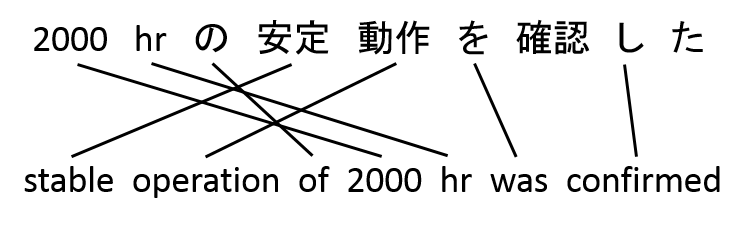}
\caption{The translation and the alignment derived from the operation sequence in Tab.~\ref{tab:osm-breakdown}.}
\label{fig:alignment}
\end{figure}

\begin{equation}
\mathtt{POP\_SRC}:\forall s\in\mathcal{V}_{src}: \begin{bmatrix}
(S)\\
()\\
()
\end{bmatrix}\rightarrow \begin{bmatrix}
(sS)\\
()\\
()
\end{bmatrix}
\label{eq:pop-rule}
\end{equation}
where $\mathcal{V}_{src}$ is the source language vocabulary. A jump from marker $X_i$ to $X_j$ is realized by replacing $P_i$ with $P_j$ in the third grammar component:

\begin{equation}
\mathtt{JMP}:\forall i,j\in \mathcal{N}: [(),(), P_i]^\mathsf{T} \rightarrow [(),(),(i P_j)]^\mathsf{T}
\label{eq:jump-rule}
\end{equation}
where $\mathcal{N}=\{k\in\mathbb{N}|k\leq n\}$ is the set of the first $n$ natural numbers for a sufficiently large $n$. The generative operations (\texttt{SET\_MARKER} and \texttt{INSERT}($t$)) insert symbols into the second component.

\begin{table}[t!]
\begin{center}
\small
\begin{tabular}{l|l}
\textbf{Derivation} & \textbf{OSNMT} \\
\hline
$[(S), (X_1), P_1]^\mathsf{T}$ & \\
$\overset{\text{Eq.~\ref{eq:pop-rule}}}{\rightarrow}[(\text{2000 } S), (X_1), P_1]^\mathsf{T}$ & \texttt{SET\_MARKER} \\
$\overset{\text{Eq.~\ref{eq:set-marker-rule}}}{\rightarrow} [(\text{2000 } S), (X_2X_1), (P_1)]^\mathsf{T}$ & 2000 \\
$\overset{\text{Eq.~\ref{eq:token-rule}}}{\rightarrow} [(\text{2000 } S), (X_2\text{ 2000 }X_1), (P_1)]^\mathsf{T}$ & \texttt{POP\_SRC} \\
$\overset{\text{Eq.~\ref{eq:pop-rule}}}{\rightarrow} \begin{bmatrix}
(\text{2000 hr } S)\\
(X_2\text{ 2000 }X_1)\\
(P_1)
\end{bmatrix}$ & hr \\
$\overset{\text{Eq.~\ref{eq:token-rule}}}{\rightarrow} \begin{bmatrix}
(\text{2000 hr } S)\\
(X_2\text{ 2000 hr }X_1)\\
(P_1)
\end{bmatrix}$ & \texttt{POP\_SRC} \\
$\overset{\text{Eq.~\ref{eq:pop-rule}}}{\rightarrow} \begin{bmatrix}
(\text{2000 hr \begin{CJK}{UTF8}{min}の\end{CJK} } S) \\
(X_2\text{ 2000 hr }X_1) \\
(P_1)
\end{bmatrix}$ & \texttt{JMP\_BWD} \\
$\overset{\text{Eq.~\ref{eq:jump-rule}}}{\rightarrow} \begin{bmatrix}
(\text{2000 hr \begin{CJK}{UTF8}{min}の\end{CJK} } S)\\
(X_2\text{ 2000 hr }X_1)\\
(\text{1 }P_2)
\end{bmatrix}$ & \texttt{SET\_MARKER} \\
$\overset{\text{Eq.~\ref{eq:set-marker-rule}}}{\rightarrow} \begin{bmatrix}
(\text{2000 hr \begin{CJK}{UTF8}{min}の\end{CJK} } S)\\
(X_3X_2\text{ 2000 hr }X_1)\\
(\text{1 }P_2)
\end{bmatrix}$ & of \\
$\overset{\text{Eq.~\ref{eq:token-rule}}}{\rightarrow} \begin{bmatrix}
(\text{2000 hr \begin{CJK}{UTF8}{min}の\end{CJK} } S) \\
(X_3\text{ of }X_2\text{ 2000 hr }X_1)\\
(\text{1 }P_2)
\end{bmatrix}$ & ... \\
... & \\
\end{tabular}
\end{center}
\caption{\label{tab:derivation} Derivation in $\mathcal{G}$ for the example of Tab.~\ref{tab:osm-breakdown}.}
\end{table}

\begin{align}
\label{eq:set-marker-rule} \mathtt{SET\_MARKER}:\forall i\in \mathcal{N}: \begin{bmatrix}
()\\
(X_i)\\
(P_i)
\end{bmatrix} \rightarrow \begin{bmatrix}
()\\
(X_{i+1}X_i)\\
(P_i)
\end{bmatrix} \\
\label{eq:token-rule} \mathtt{INSERT}:\forall i\in \mathcal{N}, t\in\mathcal{V}_{trg}: \begin{bmatrix}
()\\
(X_i)\\
(P_i)
\end{bmatrix} \rightarrow \begin{bmatrix}
()\\
(tX_i)\\
(P_i)
\end{bmatrix}
\end{align}
where $\mathcal{V}_{trg}$ is the target language vocabulary. The identity mapping $P_i\rightarrow P_i$ in the third component enforces that the write head is at marker $X_i$. We note that $\mathcal{G}$ is not only context-free but also regular in the first and third components (but not in the second component due to Eq.~\ref{eq:set-marker-rule}). Rules of the form in Eq.~\ref{eq:token-rule} are directly related to alignment links (cf.\ Fig.~\ref{fig:alignment}) as they represent the fact that target token $t$ is aligned to the last terminal symbol in the first stream. We formalize removing markers/nonterminals at the end by introducing a special nonterminal $T$ which is eventually mapped to the end-of-sentence symbol EOS:

\begin{align}
[(S),(),()]^\mathsf{T}\rightarrow [(T), (),()]^\mathsf{T} \\
[(T),(),()]^\mathsf{T}\rightarrow [(\text{EOS}), (),()]^\mathsf{T}\\
\forall i\in \mathcal{N}: [(T),  (X_i), ()]^\mathsf{T}\rightarrow[(T), (\epsilon), ()]^\mathsf{T} \\
\forall i\in \mathcal{N}: [(T),  (), (P_i)]^\mathsf{T}\rightarrow[(T), (), (\epsilon)]^\mathsf{T}
\end{align}

Tab.~\ref{tab:derivation} illustrates that there is a 1:1 correspondence between a derivation in $\mathcal{G}$ and an OSNMT operation sequence. The target-side derivation (the second component in $\mathcal{G}$) is structurally similar to a binarized version of the tree in Fig.~\ref{fig:tree}. However, we assign scores to the structure via the corresponding OSNMT sequence which does not need to obey the usual conditional independence assumptions in hierarchical SMT. Therefore, even though $\mathcal{G}$ is context-free in the second component, our scoring model for $\mathcal{G}$ is more powerful as it conditions on the OSNMT history which potentially contains context information. Note that OSNMT is deficient~\citep{ibm} as it assigns non-zero probability mass to any operation sequence, not only those with derivation in $G$. 

We further note that subword-based OSNMT can potentially represent any alignment to any target sentence as long as the alignment does not violate the 1:$n$ restriction. This is in contrast to  phrase-based SMT where reference translations often do not have a derivation in the SMT system due to coverage problems~\citep{smt-searchspace}.

\subsection{Comparison to the OSM for SMT}
\label{sec:osnmt-osm-diff}

Our OSNMT set of operations (\texttt{POP\_SRC}, \texttt{SET\_MARKER}, \texttt{JMP\_FWD}, \texttt{JMP\_BWD}, and \texttt{INSERT}($t$)) is inspired by the original OSM for SMT~\citep{osm} as it also represents the translation process as linear sequence of operations. However, there are significant differences which make OSNMT more suitable for neural models. First, OSNMT is monotone on the source side, and allows jumps on the target side. SMT-OSM operations jump in the source sentence. We argue that source side monotonicity potentially mitigates coverage issues of neural models (over- and under-translation~\citep{nmt-coverage}) as the attention can learn to scan the source sentence from left to right. Another major difference is that we use {\em markers} rather than {\em gaps}, and do not close a gap/marker after jumping to it. This is an implication of OSNMT jumps being defined on the target side since the size of a span is unknown at inference time.

\section{Training}
\label{sec:training}

\begin{algorithm}[t!]
\caption{Align2OSNMT($a$, $\mathbf{x}$, $\mathbf{y}$)}
\label{alg:align2osm}
\begin{algorithmic}[1]
\STATE{$holes\gets \{(0, \infty)\}$}
\STATE{$ops\gets\langle\rangle$}
\COMMENT{Initialize with empty list}
\STATE{$head\gets 0$}
\FOR{$i\gets 1$ \TO $|\mathbf{x}|$}
  \FORALL{$j \in \{j|a_j=i\}$}
    \STATE{$hole\_idx\gets holes.\text{find}(j)$}
    \STATE{$d\gets  hole\_idx - head$}
    \IF{$d < 0$}
      \STATE{$ops.\text{extend}(\mathtt{JMP\_BWD}.\text{repeat}(-d))$}
    \ENDIF
    \IF{$d > 0$}
      \STATE{$ops.\text{extend}(\mathtt{JMP\_FWD}.\text{repeat}(d))$}
    \ENDIF
    \STATE{$head \gets hole\_idx$}
    \STATE{$(s, t) \gets holes[head]$}
    \IF{$s \neq j$}
      \STATE{$holes.\text{append}((s, j-1))$}
      \STATE{$head\gets head + 1$}
      \STATE{$ops.\text{append}(\mathtt{SET\_MARKER})$}
    \ENDIF
    \STATE{$ops.\text{append}(\mathtt{y_j})$}
    \STATE{$holes[head]\gets (j + 1, t)$}
  \ENDFOR
  \STATE{$ops.\text{append}(\mathtt{SRC\_POP})$}
\ENDFOR
\RETURN{ops}
\end{algorithmic}
\end{algorithm}

We train our Transformer model as usual by minimising the negative log-likelihood of the target sequence. However, in contrast to plain text NMT, the target sequence is not a plain sequence of subword or word tokens but a sequence of operations. Consequently, we need to map the target sentences in the training corpus to OSNMT representations. We first run a statistical word aligner like Giza++~\citep{giza} to obtain an aligned training corpus. We delete all alignment links which violate the 1:$n$ restriction of OSNMT (cf.\ Sec.~\ref{sec:osnmt}). The alignments together with the target sentences are then used to generate the reference operation sequences for training. The algorithm for this conversion is shown in Alg.~\ref{alg:align2osm}.\footnote{A Python implementation is available at \url{https://github.com/fstahlberg/ucam-scripts/blob/master/t2t/align2osm.py}.} Note that an operation sequence represents one specific alignment, which means that the only way for an OSNMT sequence to be generated correctly is if both the word alignment and the target sentence are also correct. Thereby, the neural model learns to align and translate at the same time. However, there is spurious ambiguity as one alignment can be represented by different OSNMT sequences. For instance, simply adding a \texttt{SET\_MARKER} operation at the end of an OSNMT sequence does not change the alignment represented by it.

\section{Results}

\begin{table}[t!]
\begin{center}
\small
\begin{tabular}{l|l|r}
\textbf{Corpus} & \textbf{Language pair} & \textbf{\# Sentences} \\
\hline
Scielo & Spanish-English & 587K \\
Scielo & Portuguese-English & 513K \\
WAT & Japanese-English & 1M \\
\end{tabular}
\end{center}
\caption{\label{tab:train-set-sizes} Training set sizes.}
\end{table}

We evaluate on three language pairs: Japanese-English (ja-en), Spanish-English (es-en), and Portuguese-English (pt-en). We use the ASPEC corpus~\citep{aspec} for ja-en and the health science portion of the Scielo corpus~\citep{scielo} for es-en and pt-en. Training set sizes are summarized in Tab.~\ref{tab:train-set-sizes}. We use byte pair encoding~\citep{bpe} with 32K merge operations for all systems (joint encoding models for es-en and pt-en and separate source/target models for ja-en). We trained Transformer models~\citep{transformer}\footnote{We follow the \texttt{transformer\_base} configuration and use 6 layers, 512 hidden units, and 8 attention heads in both the encoder and decoder.} until convergence (250K steps for plain text, 350K steps for OSNMT) on a single GPU using Tensor2Tensor~\citep{t2t} after removing sentences with more than 250 tokens. Batches contain around 4K source and 4K target tokens. Transformer training is very sensitive to the batch size and the number of GPUs~\citep{t2t-training}. Therefore, we delay SGD updates~\citep{SaundersACL18} to every 8 steps to simulate 8 GPU training as recommended by~\citet{transformer}. Based on the performance on the ja-en dev set we decode the plain text systems with a beam size of 4 and OSNMT with a beam size of 8 using our SGNMT decoder~\citep{sgnmt}. We use length normalization for ja-en but not for es-en or pt-en. We report cased \texttt{multi-bleu.pl} BLEU scores on the tokenized text to be comparable with the WAT evaluation campaign on ja-en.\footnote{\url{http://lotus.kuee.kyoto-u.ac.jp/WAT/evaluation/list.php?t=2&o=4}}. 

\paragraph{Generating training alignments}

\begin{table}[t!]
\begin{center}
\small
\begin{tabular}{l|r|r|}
 & \multicolumn{2}{c|}{\textbf{BLEU}}  \\
\textbf{Method} & \textbf{es-en} & \textbf{pt-en} \\
\hline
Align on subword level & 36.7 & 38.1 \\
Convert word level alignments & 37.1 & 38.4 \\
\end{tabular}
\end{center}
\caption{\label{tab:how-to-align} Generating training alignments on the subword level.}
\end{table}

\begin{table}[t!]
\begin{center}
\small
\begin{tabular}{l|r}
\textbf{Type} & \textbf{Frequency} \\
\hline
\textbf{Valid} & \textbf{92.49\%} \\
Not enough \texttt{SRC\_POP} & 7.28\% \\
Too many \texttt{SRC\_POP} & 0.22\% \\
Write head out of range & 0.06\% \\
\end{tabular}
\end{center}
\caption{\label{tab:constraints} Frequency of invalid OSNMT sequences produced by an unconstrained decoder on the ja-en test set.}
\end{table}

As outlined in Sec.~\ref{sec:training} we use Giza++~\citep{giza} to generate alignments for training OSNMT. We experimented with two different methods to obtain alignments on the subword level. First, Giza++ can directly align the source-side subword sequences to target-side subword sequences. Alternatively, we can run Giza++ on the word level, and convert the word alignments to subword alignments in a postprocessing step by linking subwords if the words they belong to are aligned with each other. Tab.~\ref{tab:how-to-align} compares both methods and shows that converting word alignments is marginally better. Thus, we use this method in all other experiments.

\paragraph{Constrained beam search}

Unconstrained neural decoding can yield invalid OSNMT sequences. For example, the \texttt{JMP\_FWD} and \texttt{JMP\_BWD} operations are undefined if the write head is currently at the position of the last or first marker, respectively. The number of \texttt{SRC\_POP} operations must be equal to the number of source tokens in order for the read head to scan the entire source sentence. Therefore, we constrain these operations during decoding. We have implemented the constraints in our publicly available SGNMT decoding platform~\citep{sgnmt}. However, these constraints are only needed for a small fraction of the sentences. Tab.~\ref{tab:constraints} shows that even unconstrained decoding yields valid OSNMT sequences in 92.49\% of the cases.

\begin{table}[t!]
\begin{center}
\small
\begin{tabular}{l|r|r|r|r|}
  & \multicolumn{4}{c|}{\textbf{BLEU}} \\
  & \textbf{es-en} & \textbf{pt-en} & \multicolumn{2}{c|}{\textbf{ja-en}} \\
\textbf{Representation} & & & \textbf{dev} & \textbf{test} \\
\hline
Plain & 37.6 & 37.5 & 28.3 & 28.1 \\
OSNMT & 37.1 & 38.4 & 28.1 & 28.8 \\
\end{tabular}
\end{center}
\caption{\label{tab:plain-vs-osnmt} Comparison between plain text and OSNMT on Spanish-English (es-en), Portuguese-English (pt-en), and Japanese-English (ja-en).}
\end{table}

\begin{table*}[t!]
\begin{center}
\footnotesize
\begin{tabular}{|p{15cm}|}
\hline
\begin{center}\includegraphics[scale=0.75]{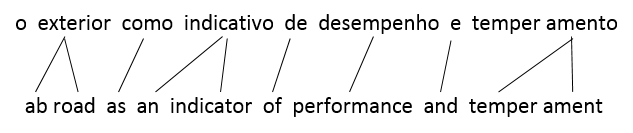}\end{center} \\
\textbf{Operation sequence}: \texttt{SRC\_POP} ab road\_ \texttt{SRC\_POP} as\_ \texttt{SRC\_POP} an\_ indicator\_ \texttt{SRC\_POP} of\_ \texttt{SRC\_POP} performance\_ \texttt{SRC\_POP} and\_ \texttt{SRC\_POP} \texttt{SRC\_POP} temper ament\_ \texttt{SRC\_POP} \\
\textbf{Reference}:  the body shape as an indicative of performance and temperament \\
\hline
\begin{center}\includegraphics[scale=0.75]{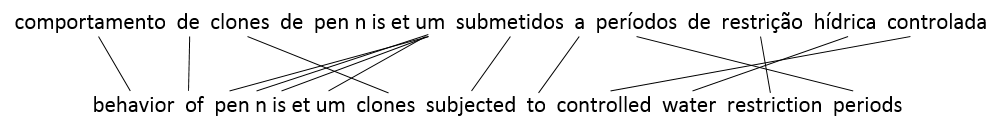}\end{center} \\
\textbf{Operation sequence}: behavior\_ \texttt{SRC\_POP} of\_ \texttt{SRC\_POP} \texttt{SET\_MARKER} clones\_ \texttt{SRC\_POP} \texttt{SRC\_POP} \texttt{SRC\_POP} \texttt{SRC\_POP} \texttt{SRC\_POP} \texttt{JMP\_BWD} pen n is et um\_ \texttt{SRC\_POP} \texttt{JMP\_FWD} subjected\_ \texttt{SRC\_POP} to\_ \texttt{SRC\_POP} \texttt{SET\_MARKER} periods\_ \texttt{SRC\_POP} \texttt{SRC\_POP} \texttt{JMP\_BWD} \texttt{SET\_MARKER} restriction\_ \texttt{SRC\_POP} \texttt{JMP\_BWD} \texttt{SET\_MARKER} water\_ \texttt{SRC\_POP} \texttt{JMP\_BWD} controlled\_ \texttt{SRC\_POP} \\
\textbf{Reference}:  response of pennisetum clons to periods of controlled hidric restriction \\
\hline
\begin{center}\includegraphics[scale=0.75]{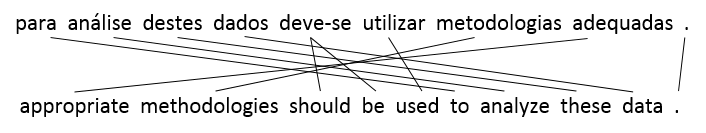}\end{center} \\
\textbf{Operation sequence}: \texttt{SET\_MARKER} to\_ \texttt{SRC\_POP} analyze\_ \texttt{SRC\_POP} these\_ \texttt{SRC\_POP} data\_ \texttt{SRC\_POP} \texttt{JMP\_BWD} \texttt{SET\_MARKER} should\_ be\_ \texttt{SRC\_POP} used\_ \texttt{SRC\_POP} \texttt{JMP\_BWD} \texttt{SET\_MARKER} methodologies\_ \texttt{SRC\_POP}> \texttt{JMP\_BWD} appropriate\_ \texttt{SRC\_POP} \texttt{JMP\_FWD} \texttt{JMP\_FWD} \texttt{JMP\_FWD} .\_ \texttt{SRC\_POP} \\
\textbf{Reference}:  to analyze these data suitable methods should be used . \\
\hline
\end{tabular}
\end{center}
\caption{\label{tab:pten-alignments} Examples of Portuguese-English translations together with their (subword-)alignments induced by the operation sequence. Alignment links from source words consisting of multiple subwords were mapped to the final subword in the training data, visible for `temperamento' and `pennisetum'.}
\end{table*}

\paragraph{Comparison with plain text NMT}

Tab.~\ref{tab:plain-vs-osnmt} compares our OSNMT systems with standard plain text models on all three language pairs. OSNMT performs better on the pt-en and ja-en test sets, but slightly worse on es-en. We think that more engineering work such as optimizing the set of operations or improving the training alignments could lead to more consistent gains from using OSNMT. However, we leave this to future work since the main motivation for this paper is explainable NMT and not primarily improving translation quality.

\paragraph{Alignment quality}
\label{sec:pten_alignment}

Tab.~\ref{tab:pten-alignments} contains example translations and subword-alignments generated from our Portuguese-English OSNMT model. Alignment links from source words consisting of multiple subwords are mapped to the final subword, visible for the words `temperamento' in the first example and `pennisetum' in the second one. The length of the operation sequences increases with alignment complexity as operation sequences for monotone alignments consist only of \texttt{INSERT}($t$) and \texttt{SRC\_POP} operations (example 1). However, even complex mappings are captured very well by OSNMT as demonstrated by the third example. Note that OSNMT can represent long-range reorderings very efficiently: the movement from `para' in the first position to `to' in the tenth position is simply achieved by starting the operation sequence with `\texttt{SET\_MARKER} to' and a \texttt{JMP\_BWD} operation later. The first example in particular demonstrates the usefulness of such alignments as the wrong lexical choice (`abroad' rather than `body shape') can be traced back to the source word `exterior`.

\begin{figure}[!t]
\centering
\includegraphics[width=1\linewidth]{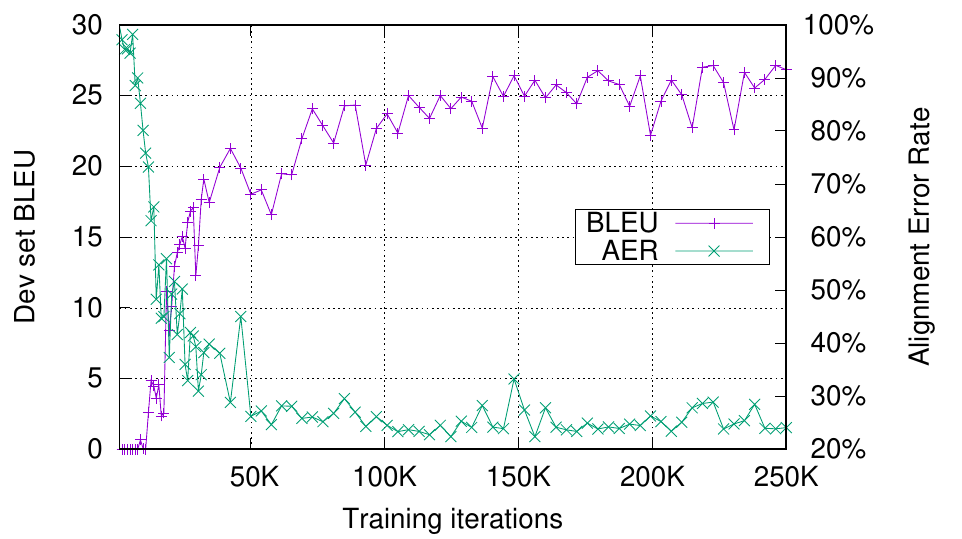}
\caption{AER and BLEU training curves for OSNMT on the Japanese-English dev set.}
\label{fig:aer-bleu}
\end{figure}

\begin{table*}[t!]
\begin{center}
\small
\begin{tabular}{l|l|r|r|}
 &  &  \multicolumn{2}{c|}{\textbf{AER (in \%)}} \\
\textbf{Representation} & \textbf{Alignment extraction} & \textbf{dev} & \textbf{test} \\
\hline
Plain & LSTM forced decoding & 63.9 & 63.7 \\
Plain & LSTM forced decoding with supervised attention \citep[{\em Cross Entropy} loss]{supervised-att3} & 54.9 & 54.7 \\
OSNMT & OSNMT & 24.2 & 21.5 \\
\end{tabular}
\end{center}
\caption{\label{tab:rnn-vs-osnmt} Comparison between OSNMT and using the attention matrix from forced decoding with a recurrent model.}
\end{table*}

\begin{figure*}
    \centering
    \begin{subfigure}[b]{0.45\textwidth}
        \includegraphics[width=\textwidth]{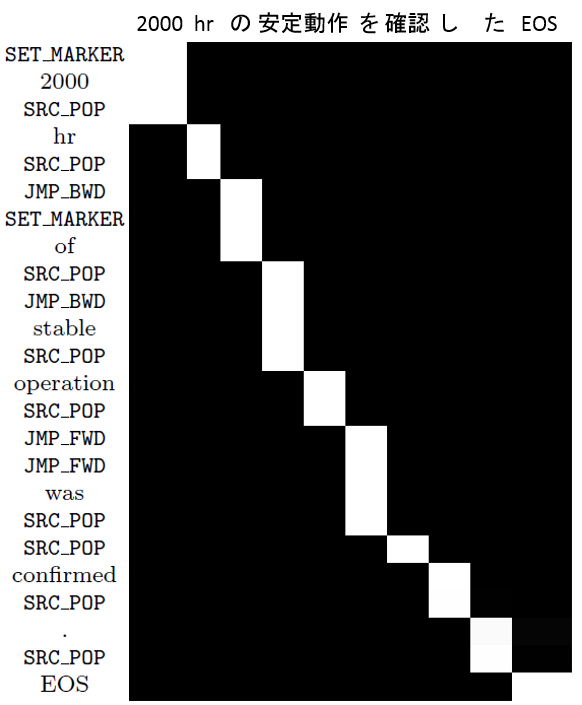}
        \caption{Layer 4, head 1; attending to the source side read head.}
        \label{fig:encdec1}
    \end{subfigure}
    \begin{subfigure}[b]{0.45\textwidth}
        \includegraphics[width=\textwidth]{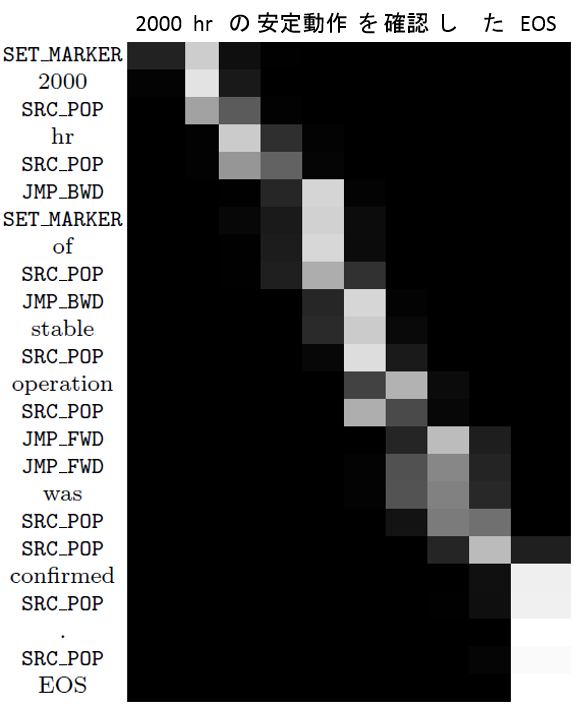}
        \caption{Layer 2, head 3; attending to the right trigram context of the read head.}
        \label{fig:encdec2}
    \end{subfigure}
    \caption{Encoder-decoder attention weights.}\label{fig:encdec}
\end{figure*}

For a qualitative assessment of the alignments produced by OSNMT we ran Giza++ to align the generated translations to the source sentences, enforced the $1$:$n$ restriction of OSNMT, and used the resulting alignments as reference for computing the alignment error rate~\citep[AER]{giza}. Fig.~\ref{fig:aer-bleu} shows that as training proceeds, OSNMT learns to both produce high quality translations (increasing BLEU score) and accurate alignments (decreasing AER).

As mentioned in the introduction, a light-weight way to extract $1$:$n$ alignments from a vanilla attentional LSTM-based seq2seq model is to take the maximum over attention weights for each target token. This is possible because, unlike the Transformer, LSTM-based models usually only have a single soft attention matrix. However, in our experiments, LSTM-based NMT was more than 4.5 BLEU points worse than the Transformer on Japanese-English. Therefore, to compare AERs under comparable BLEU scores, we used the LSTM-based models in forced decoding mode on the output of our plain text Transformer model from Tab.~\ref{tab:plain-vs-osnmt}. We trained two different LSTM models: one standard model by optimizing the likelihood of the training set, and a second one with supervised attention following \citet{supervised-att3}. Tab.~\ref{tab:rnn-vs-osnmt} shows that the supervised attention loss of \citet{supervised-att3} improves the AER of the LSTM model. However, OSNMT is able to produce much better alignments since it generates the alignment along with the translation in a single decoding run.

\paragraph{OSNMT sequences contain target words in source sentence order}

An OSNMT sequence can be seen as a sequence of target words in source sentence order, interspersed with instructions on how to put them together to form a fluent target sentence. For example, if we strip out all \texttt{SRC\_POP}, \texttt{SET\_MARKER}, \texttt{JMP\_FWD}, and \texttt{JMP\_BWD} operations in the OSNMT sequence in the second example of Tab.~\ref{tab:pten-alignments} we get:

\begin{quote}
behavior of clones pennisetum subjected to periods restriction water controlled
\end{quote}

The word-by-word translation back to Portugese is:

\begin{quote}
comportamento de clones pennisetum submetidos a períodos restrição hídrica controlada
\end{quote}

This restores the original source sentence (cf.\ Tab.~\ref{tab:pten-alignments}) up to unaligned source words. Therefore, we can view the operations for controlling the write head (\texttt{SET\_MARKER}, \texttt{JMP\_FWD}, and \texttt{JMP\_BWD}) as reordering instructions for the target words which appear in source sentence word order within the OSNMT sequence.

\paragraph{Role of multi-head attention}
\label{sec:att}

In this paper, we use a standard seq2seq model (the Transformer architecture~\citep{transformer}) to generate OSNMT sequences from the source sentence. This means that our neural model is representation-agnostic: we do not explicitly incorporate the notion of read and write heads into the neural architecture. In particular, neither in training nor in decoding do we explicitly bias the Transformer's attention layers towards consistency with the alignment represented by the OSNMT sequence. Our Transformer model has 48 encoder-decoder attention matrices due to multi-head attention (8 heads in each of the 6 layers). We have found that many of these attention matrices have strong and interpretable links to the translation process represented by the OSNMT sequence. For example, Fig.~\ref{fig:encdec1} shows that the first head in layer 4 follows the source-side read head position very closely: at each \texttt{SRC\_POP} operation the attention shifts by one to the next source token. Other attention heads have learned to take other responsibilities. For instance, head 3 in layer 2 (Fig.~\ref{fig:encdec2}) attends to the trigram right of the source head. 

\section{Related Work}

Explainable and interpretable machine learning is attracting more and more attention in the research community~\citep{xai-blackbox,xai-survey}, particularly in the context of natural language processing~\citep{visual-rnn,visual-nlp,causal-seq2seq,visual-nmt,seq2seq-pathologies}. These approaches aim to explain (the predictions of) an existing model. In contrast, we change the target representation such that the generated sequences themselves convey important information about the translation process such as the word alignments.

Despite considerable consensus about the importance of word alignments in practice~\citep{nmt-challenges}, e.g.\ to enforce constraints on the output~\citep{terminology-constraints} or to preserve text formatting, introducing explicit alignment information to NMT is still an open research problem. Word alignments have been used as supervision signal for the NMT attention model~\citep{supervised-att1,supervised-att2,supervised-att3,supervised-att-rwth}. \citet{ibm-like-attention} showed how to reintroduce concepts known from traditional statistical alignment models~\citep{ibm} like fertility and agreement over translation direction to NMT. Some approaches to simultaneous translation explicitly control for reading source tokens and writing target tokens and thereby generate monotonic alignments on the segment level~\citep{simtrans-blunsom,simtrans-blunsom-noisy-channel,simtrans-neubig}. \citet{alignment-based-nmt} used separate alignment and lexical models and thus were able to hypothesize explicit alignment links during decoding. While our motivation is very similar to \citet{alignment-based-nmt}, our approach is very different as we represent the alignment as operation sequence, and we do not use separate models for reordering and lexical translation.

The operation sequence model for SMT~\citep{osm,osm-journal} has been used in a number of MT evaluation systems~\citep{eval-uedin-wmt14,eval-rwth-iwslt,eval-qcri-iwslt} and for post-editing~\citep{osm-post-editing}, often in combination with a phrase-based model. The main difference to our OSNMT is that we have adapted the set of operations for neural models and are able to use it as stand-alone system, and not on top of a phrase-based system.

Our operation sequence model has some similarities with transition-based models used in other areas of NLP~\citep{transition-based-parsing-rnn1,transition-based-parsing-rnn2,morph-hard-alignment}. In particular, our \texttt{POP\_SRC} operation is very similar to the {\em step} action of the hard alignment model of \citet{morph-hard-alignment}. However, \citet{morph-hard-alignment} investigated monotonic alignments for morphological inflections whereas we use a larger operation/action set to model complex word reorderings in machine translation.

\section{Conclusion}

We have presented a way to use standard seq2seq models to generate a translation together with an alignment as linear sequence of operations. This greatly improves the interpretability of the model output as it establishes explicit alignment links between source and target tokens. However, the neural architecture we used in this paper is representation-agnostic, i.e.\ we did not explicitly incorporate the alignments induced by an operation sequence into the neural model. For future work we are planning to adapt the Transformer model, for example by using positional embeddings of the source read head and the target write head in the Transformer attention layers.

\section*{Acknowledgments}

This work was supported in part by the U.K. Engineering and Physical Sciences Research Council (EPSRC grant EP/L027623/1). We thank Joanna Stadnik who produced the recurrent translation and alignment models during her 4th year project.


\bibliography{refs}
\bibliographystyle{acl_natbib_nourl}

\end{document}